\documentclass[10pt,twocolumn,letterpaper]{article}

\usepackage[pagenumbers]{cvpr} 

\usepackage{graphicx}
\usepackage{amsmath}
\usepackage{amssymb}
\usepackage[normalem]{ulem}
\usepackage{algorithm}
\usepackage{algorithmic}
\usepackage{bm}
\usepackage{soul}
\usepackage{multirow} 
\usepackage{makecell}
\usepackage{times}
\usepackage{epsfig}
\usepackage{graphicx}
\usepackage{dblfloatfix}    
\usepackage{xcolor} 
\usepackage[flushleft]{threeparttable} 
\usepackage{multirow} 
\usepackage{wrapfig,lipsum,booktabs} 

\usepackage{amsfonts}       
\usepackage{nicefrac}       
\usepackage{microtype}      
\usepackage{caption}
\usepackage{mathtools}
\usepackage{amsmath, bm}
\usepackage{amsfonts}
\usepackage{algorithm,algorithmic}

\usepackage{listings}

\definecolor{codegreen}{rgb}{0,0.6,0}
\definecolor{codegray}{rgb}{0.5,0.5,0.5}
\definecolor{codepurple}{rgb}{0.58,0,0.82}
\definecolor{backcolour}{rgb}{0.95,0.95,0.92}

\lstdefinestyle{mystyle}{
    backgroundcolor=\color{backcolour},   
    commentstyle=\color{codegreen},
    keywordstyle=\color{magenta},
    numberstyle=\tiny\color{codegray},
    stringstyle=\color{codepurple},
    basicstyle=\ttfamily\footnotesize,
    breakatwhitespace=false,         
    breaklines=true,                 
    captionpos=b,                    
    keepspaces=true,                 
    numbers=left,                    
    numbersep=5pt,                  
    showspaces=false,                
    showstringspaces=false,
    showtabs=false,                  
    tabsize=2
}

\newcommand{\va}{\mathbf{a}}
\newcommand{\vw}{\mathbf{w}}
\newcommand{\vx}{\mathbf{x}}
\newcommand{\vy}{\mathbf{y}}

\newcommand{\vpi}{{\bm{\pi}}}

\let\oldtextbf=\textbf
\renewcommand\textbf[1]{{\boldmath\oldtextbf{#1}}}

\usepackage[pagebackref,breaklinks,colorlinks]{hyperref}

\usepackage[capitalize]{cleveref}
\crefname{section}{Sec.}{Secs.}
\Crefname{section}{Section}{Sections}
\Crefname{table}{Table}{Tables}
\crefname{table}{Tab.}{Tabs.}

\def\cvprPaperID{***} 
\def\confName{CVPR}
\def\confYear{2022}

\begin{document}


\newcommand{\PaperTitle}{FBNetV5: Neural Architecture Search for Multiple Tasks in One Run}
\newcommand{\FrameworkNameFull}{FBNetV5}
\newcommand{\FrameworkName}{FBNetV5}

\title{\PaperTitle}

\author{Bichen Wu$^{1}$\thanks{Equal contribution. $\dagger$ Work done while interning at Meta Reality Labs.}, Chaojian Li$^{2*\dagger}$, Hang Zhang$^1$, Xiaoliang Dai$^1$, Peizhao Zhang$^1$, \\
Matthew Yu$^1$, Jialiang Wang$^1$, Yingyan Lin$^2$, Peter Vajda$^1$\\
$^1$Meta Reality Labs, $^2$Rice University\\
{\tt\small \{wbc,zhanghang,xiaoliangdai,stzpz,mattcyu,jialiangw,vajdap\}@fb.com} \\
{\tt\small \{cl114,yingyan.lin\}@rice.edu}
}
\maketitle

\begin{abstract}
Neural Architecture Search (NAS) has been widely adopted to design accurate and efficient image classification models. However, applying NAS to a new computer vision task still requires a huge amount of effort. This is because 1) previous NAS research has been over-prioritized on image classification while largely ignoring other tasks; 2) many NAS works focus on optimizing task-specific components that cannot be favorably transferred to other tasks; and 3) existing NAS methods are typically designed to be ``proxyless'' and require significant effort to be integrated with each new task's training pipelines. To tackle these challenges, we propose {\FrameworkName}, a NAS framework that can search for neural architectures for a variety of vision tasks with much reduced computational cost and human effort. Specifically, we design 1) a search space that is simple yet inclusive and transferable; 2) a multitask search process that is disentangled with target tasks' training pipeline; and 3) an algorithm to simultaneously search for architectures for multiple tasks with a computational cost agnostic to the number of tasks. We evaluate the proposed {\FrameworkName} targeting three fundamental vision tasks -- image classification, object detection, and semantic segmentation. Models searched by {\FrameworkName} in a \textbf{single} run of search have outperformed the previous state-of-the-art in all the three tasks: image classification (e.g., $\uparrow$1.3\% ImageNet top-1 accuracy under the same FLOPs as compared to FBNetV3), semantic segmentation (e.g., $\uparrow$1.8\% higher ADE20K \textit{val.} mIoU than SegFormer with 3.6$\times$ fewer FLOPs), and object detection (e.g., $\uparrow$1.1\% COCO \textit{val.} mAP with 1.2$\times$ fewer FLOPs as compared to YOLOX).
\end{abstract}

\begin{figure}[t]
  \centering
  \includegraphics[width=0.99\linewidth]{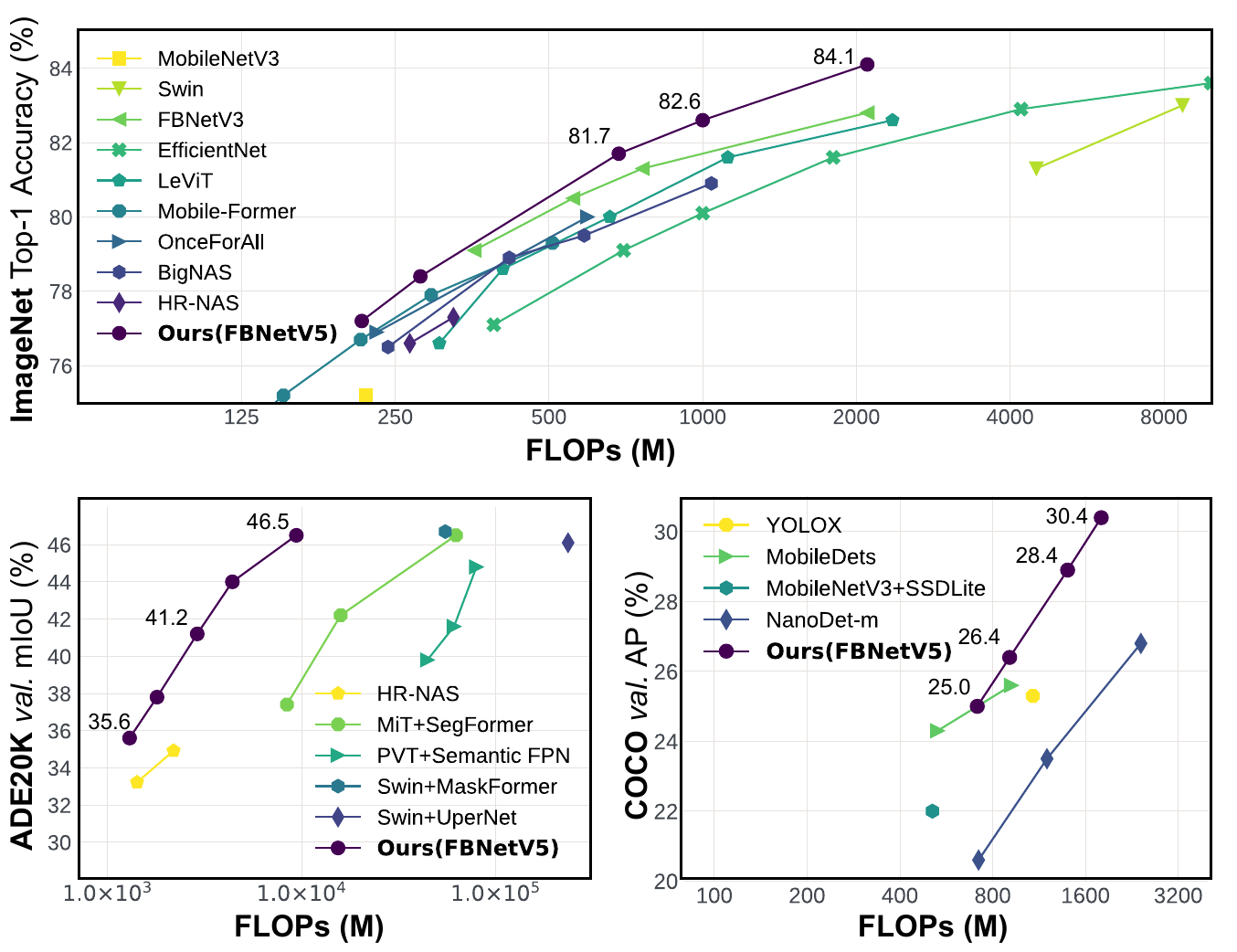}
\vspace{-1em}
\caption{The architectures simultaneously searched in a single run of {\FrameworkName} outperforms the SotA performance in three tasks: ImageNet~\cite{deng2009imagenet} image classification, ADE20K~\cite{zhou2017scene} semantic segmentation, and COCO~\cite{lin2014microsoft} object detection.}
  \vspace{-1.5em}
\label{fig:performance}
\end{figure}

\vspace{-1.5em}
\section{Introduction}

Recent breakthroughs in deep neural networks (DNNs) have fueled a growing demand for deploying DNNs in perception systems for a wide range of computer vision (CV) applications that are powered by various fundamental CV tasks, including classification, object detection, and semantic segmentation. To develop real-world DNN based perception systems, the neural architecture design is among the most important factors that determine the achievable task performance and efficiency. Nevertheless, designing neural architectures for different applications is challenging due to its prohibitive computational cost, intractable design space~\cite{radosavovic2020designing,dong2020bench,Ci_2021_ICCV}, diverse application-driven deployment requirements~\cite{wu2019fbnet,li2021searching,xiong2020mobiledets}, and so on. 

To tackle the aforementioned challenges, the CV community has been exploring neural architecture search (NAS) to design DNNs for CV tasks. In general, the expectations for NAS are two-fold: First, to build better neural architectures with stronger performance and higher efficiency; and second, to automate the design process in order to reduce the  human effort and computational cost for DNN design. While the former ensures effective real-world solutions, the latter is critical to facilitate the fast development of DNNs to more applications. Looking back at the progress of recent years, it is fair to say that NAS has met the first expectation in advancing the frontiers of accuracy and efficiency, especially for image classification tasks. However, existing NAS methods still fall short of meeting the second expectation.

The reasons for the above limitation include the following. \underline{First}, over the years the NAS community has been over fixated on benchmarking NAS methods on image classification tasks, driven by the commonly believed assumption that the best models for image classification are also the best backbones for other tasks. However, this assumption is not always true~\cite{xiong2020mobiledets,du2020spinenet,chen2019detnas,zhang2021dcnas}, and often leads to suboptimal architectures for many non-classification tasks. \underline{Second}, many existing NAS works focus on optimizing task-specific components that are not transferable or favorable to other tasks. For example, \cite{shaw2019squeezenas} only searches for the encoder part within the encoder-decoder structure of segmentation tasks, while the optimal encoder is coupled with the decoder designs. \cite{ghiasi2019fpn} is customized to RetinaNet~\cite{lin2017focal} in object detection tasks.
As a result, NAS advances made for one task do not necessarily favor other tasks or help reduce the design effort. \underline{Finally}, a popular belief in current NAS practice is that it is better for NAS to be ``proxyless'' and a NAS method should be integrated into the target tasks' training pipeline for directly optimizing the corresponding architectures based on the training losses of each target task~\cite{cai2019once,cai2018proxylessnas,yu2020bignas}. 
However, this makes NAS unscalable when dealing with many new tasks, since adding each new task would require nontrivial efforts to integrate the NAS techniques into the existing training pipeline of the target task. In particular, many popular NAS methods conduct search by training a supernet~\cite{yu2020bignas,cai2019once,wang2021alphanet}, adding dedicated cost regularization to the loss function~\cite{ding2021hr}, adopting special initialization~\cite{yu2020bignas}, and so on. These techniques often heavily interfere with the target task's training process and thus requires much engineering effort to re-tune the hyperparameters to achieve the desired performance.

In this work, we propose \textbf{\FrameworkName}, a NAS framework, that can simultaneously search for backbone topologies for multiple tasks in a single run of search. As a proof of concept, we target three fundamental computer vision tasks -- image classification, object detection, and semantic segmentation. Starting from a state-of-the-art image classification model, \ie, FBNetV3 \cite{Dai2020FBNetV3JA}, we construct a \textit{supernet} consisting of parallel paths with multiple resolutions, similar to HRNet~\cite{wang2020deep,ding2021hr}. Based on the supernet, {\FrameworkName} searches for the optimal topology for each target task by parameterizing a set of binary masks indicating whether to keep or drop a building block in the supernet. To disentangle the search process from the target tasks' training pipeline, we conduct search by training the supernet on a proxy multitask dataset with classification, object detection, and semantic segmentation labels. Following \cite{ghiasi2021multi}, the dataset is based on ImageNet, with detection and segmentation labels generated by pretrained open-source models. To make the computational cost and hyper-parameter tuning effort agnostic to the number of tasks, we propose a \textbf{supernet training algorithm} that \textit{simultaneously search} for task architectures \textit{in one run}. 
After the supernet training, we individually train the searched task-specific architectures to uncover their performance.

Excitingly, in addition to requiring reduced computational cost and human effort, extensive experiments show that {\FrameworkName} produces compact models that can achieve SotA performance on all three target tasks. On ImageNet \cite{deng2009imagenet} classification, our model achieved  $1.3$\% higher top-1 accuracy under the same FLOPs as compared to FBNetV3 \cite{Dai2020FBNetV3JA}; on ADE20K~\cite{zhou2017scene} semantic segmentation, our model achieved $1.8$\% higher mIoU than SegFormer \cite{xie2021segformer} with 3.6$\times$ fewer FLOPs; on COCO \cite{lin2014microsoft} object detection, our model achieved 1.1\% higher mAP with 1.2$\times$ fewer FLOPs compared to YOLOX \cite{ge2021yolox}. 
It is worth noting that all our well-performing architectures are searched simultaneously in \textit{a single run}, yet they beat the SotA neural architectures that are delicately searched or designed \textit{for each task}. 

\begin{figure*}[t]
  \centering
  \includegraphics[width=1.0\linewidth]{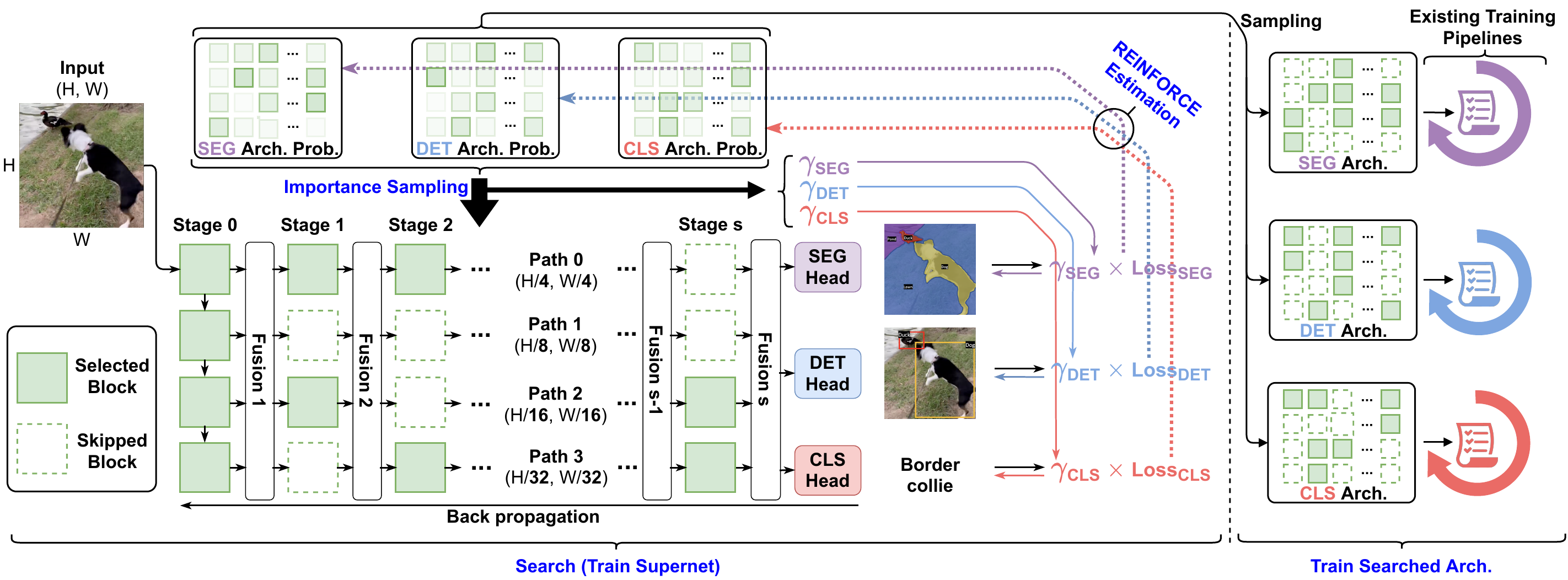}
\vspace{-1.5em}
\caption{Overview of {\FrameworkName}. We search backbone topologies for multiple tasks by training a supernet once on a multitask dataset. Each task has its own architecture distribution from which we sample task-specific architectures and train them using the existing training pipeline of the target tasks. Supernet configurations in Appendix \ref{sec:block_config}. Fusion module details in Appendix~\ref{sec:fusion_details}. Search process in Algorithm \ref{alg:multitask_supernet_is_reinforce}. 
}
  \vspace{-1.em}
\label{fig:overview}
\end{figure*}

\section{Related Works}
\textbf{Neural Architecture Search for Efficient DNNs.}
Various NAS methods have been developed to design efficient DNNs, aiming to 1) achieve boosted accuracy vs. efficiency trade-offs~\cite{he2016deep,sandler2018mobilenetv2,hu2018squeeze} and 2) automate the design process to reduce human effort and computational cost. Early NAS works mostly adopt reinforcement learning~\cite{zoph2016neural,tan2019mnasnet} or evolutionary search algorithms~\cite{real2017large} which require substantial resources. To reduce the search cost, differentiable NAS~\cite{wu2019fbnet,wan2020fbnetv2, cai2018proxylessnas, chen2019progressive,liu2018darts} was developed to differentiably update the weights and architectures. Recently, to deliver multiple neural architectures meeting different cost constraints,~\cite{cai2019once,yu2020bignas} propose to jointly train all the sub-networks in a weight-sharing supernet and then locate the optimal architectures under different cost constraints without re-training or fine-tuning. However, unlike our work, all the works above focus on a single task, mostly image classification, and they do not reduce the effort of designing architectures for other tasks. 

\textbf{Task-aware Neural Architecture Design.} 
To facilitate designing optimal DNNs for various tasks, recent works ~\cite{howard2019searching,liu2021swin,wang2020deep} propose to design general architecture backbones for different CV tasks.
In parallel, 
with the belief that each CV task requires its own unique architecture to achieve the task-specific optimal accuracy vs. efficiency trade-off,~\cite{shaw2019squeezenas,chen2019fasterseg,ghiasi2019fpn,lin2017focal,tan2020efficientdet} develop dedicated search spaces for different CV tasks, from which they search for task-aware DNN architectures. However, these existing methods mostly focus on optimizing task-specific components of which the advantages are not transferable to other tasks.
Recent works \cite{ding2021hr,cheng2020scalenas} begin to focus on designing networks for multiple tasks in a unified search space and has shown promising results. However, they are designed to be ``proxyless'' and the search process needs to be integrated to downstream tasks's training pipeline. This makes it less scalable to add new tasks, since it requires non-trivial engineering effort and compute cost to integrate NAS to the existing training pipeline of a target task. Our work bypasses this by using a disentangled search process, and we conduct search for multiple tasks in one run. This is computationally efficient and allows us to utilize  target tasks' existing training pipelines with no extra efforts.
\section{Method}
\label{sec:method}
In this section, we present our proposed {\FrameworkName} framework that aims to reduce the computational cost and human effort required by NAS for multiple tasks. {\FrameworkName} contains three key components: 1) A simple yet inclusive and transferable search space (Section \ref{sec:space}); 2) A search process equipped with a multitask learning proxy to disentangle NAS from target tasks' training pipelines (Section \ref{sec:search}); and 3) a search algorithm to simultaneously produce architectures for multiple tasks at a constant computational cost agnostic to the number of target tasks (Section \ref{sec:search-algorithm}).

\subsection{Search Space}
\label{sec:space}
To search for architectures for multiple tasks, we design the search space to meet three standards: 1) \textbf{Simple and elegant}: we favor simple search space over complicated ones; 2) \textbf{Inclusive}: the search space should include strong architectures for all target tasks; and 3) \textbf{Transferable}: the searched architectures should be useful not only for one model, but also transferable to a family of models.

Inspired by HRNet\cite{wang2020deep, ding2021hr}, we extend a SotA classification model, FBNetV3~\cite{Dai2020FBNetV3JA}, to a supernet with parallel paths and multiple stages. Each path has a different resolution while blocks on the same path have the same resolution. This is shown in Figure~\ref{fig:overview} (bottom-left). We divide an FBNetV3 into 4 partitions along the depth dimension, each partition outputs a feature map with a resolution down-sampled by $4$, $8$, $16$, and $32$ times, respectively. \texttt{Stage 0} of the supernet is essentially the FBNetV3 model. For following stages, we use the last 2 layers of each partition to construct a block per stage. During inference, we first compute \texttt{Stage 0} of the supernet, and then compute the remaining blocks by topological order. Similar to ~\cite{wang2020deep}, we insert (lightweight) fusion modules  (see Appendix~\ref{sec:fusion_details}) between stages to fuse information from different paths (resolutions). A block-wise model configuration of the supernet can be found in Appendix \ref{sec:block_config}. 

The aforementioned supernet contains blocks with varying significance to different tasks. By conventional wisdom, a classification architecture may only need blocks on the low-resolution paths, while segmentation or object detection would favor blocks with a higher resolution. Based on this, we search for network topologies, \ie, which blocks to select or skip for different tasks. Formally, for a supernet with $P$ paths, $S$ stages, and $B = S \times P$ blocks, a candidate architecture can be characterized by a binary vector $\va \in \{0, 1\}^B$, where $\va_i=1$ means to select block-$i$ and  $\va_i=0$ means to skip and remove the corresponding connections from and to this block. More details about the implementation of fusion modules with skipped blocks are provided in Appendix~\ref{sec:fusion_details}.

We believe that this search space is \textbf{simple and elegant}. It only contains binary choices for each of the $B$ blocks. This is much simpler than other search space design that considers how to mix different types of operators (convolutions and transformers) together, or how to wire operators with complicated connections. Furthermore, the search space is  \textbf{inclusive}. As a sanity check, the search space include most of the mainstream network topologies for CV tasks, \eg, 1) the simple linear topology for most of the classification models, 2) the U-Net \cite{ronneberger2015u} and PANet \cite{liu2018path} topology for semantic segmentation, and 3) the Feature-Pyramid Networks (FPN) \cite{lin2017feature} and BiFPN \cite{tan2020efficientdet} for object detection, as illustrated in Appendix~\ref{sec:special_cases}. The searched architecture topology is \textbf{transferable}. FBNet contains a series of models from small to large. We conduct search on a FBNet-A based supernet, and the topology can be transferred to other models. It is worth noting that transferring topology to models of different sizes, depths, and resolutions is also a common practice adopted by works such as FPN \cite{lin2017feature} and BiFPN \cite{tan2020efficientdet}).

\subsection{Disentangled Search Process}
\label{sec:search}
A popular belief is that NAS should be proxyless and the search process should be integrated into each target task's training pipeline for achieving better results. However, implementing and integrating the search process to each target task's pipeline can require significant engineering effort. Moreover, many NAS techniques heavily interfere with the target task's training and thus requires much engineering effort to re-tune the hyperparameters.

To avoid the above limitations, we design a search process that is disentangled with target tasks' training pipeline. Specifically, we conduct search by training a supernet on a multitask dataset where each image is annotated with labels from all target tasks. Following \cite{wu2019fbnet}, the supernet training jointly optimize the model weights and more importantly, task-specific architecture distributions (\eg, the \texttt{SEG}, \texttt{DET}, and \texttt{CLS} Arch. Prob. in Figure~\ref{fig:overview}). The goal of the search process is to obtain a task-specific architecture distribution from which we can sample architectures for the target tasks. The searched models can then be trained using the existing training pipeline of the target tasks without the necessity of implementing the search process into the tasks' training pipeline or re-tune the existing hyper-parameters. The search process is shown in Figure \ref{fig:overview}.

As there is no large-scale multitask dataset publicly available, we follow \cite{ghiasi2021multi} to construct a pseudo-labeled dataset based on ImageNet. Specifically, we use 1) original ImageNet labels for classification, 2) open-source CenterNet2 \cite{zhou2021probablistic} pretrained on the COCO object detection dataset to generate pseudo detection labels, and 3) open-source MaskFormer \cite{cheng2021maskformer} pretrained on the COCO-stuff semantic segmentation dataset (171 classes) to generate pixel-wise segmentation labels. In addition, we follow \cite{ghiasi2021multi} to filter out object detection results with a confidence lower than 0.5, and set segmentation predictions whose maximum probability lower than 0.5 to be the ``don't-care'' category. As such, this dataset can easily extend to include more tasks by using open-source pretrained models to generate task-specific pseudo labels.

\subsection{Search Algorithm}
\label{sec:search-algorithm}
Our search algorithm is based on the differentiable neural architecture search \cite{liu2018darts, wu2019fbnet, wan2020fbnetv2} for low computational cost compared with other methods, such as sampling-based methods \cite{Dai2020FBNetV3JA,Tan_2019_CVPR}. For multiple tasks, a simple idea is to apply the conventional single-task NAS (Algorithm~\ref{alg:singletask_supernet}) $T$ times for each task.  To make this more scalable, we derive a novel search algorithm with a constant computational cost agnostic to the number of tasks  (Algorithm~\ref{alg:multitask_supernet_is_reinforce}). For better clarity, We introduce the derivation of the search algorithm in four steps corresponding to Algorithm~\ref{alg:singletask_supernet}, \ref{alg:multitask_supernet}, \ref{alg:multitask_supernet_is}, and \ref{alg:multitask_supernet_is_reinforce}, respectively. We summarize and compare the four search algorithms at each step in Table \ref{tab:algs_comparision}. We visualize Algorithm \ref{alg:multitask_supernet_is_reinforce} in Figure \ref{fig:overview}. 

\begin{table}[t]
\caption{Summary of the differentiable NAS algorithms. $T$ represents the number of tasks.}
\centering
  \resizebox{1.0\linewidth}{!}
  {
    \begin{tabular}{c||c|cc}
    \toprule
    \multirow{3}{*}{\textbf{Search Algorithms}} & \multirow{3}{*}{\textbf{\#Tasks to Handle}} &
    \multicolumn{2}{c}{\textbf{Search Cost}} \\
    \addlinespace[-0.2em]
    \cmidrule{3-4}
    \addlinespace[-0.25em]
     & & \textbf{\#Forward} & \textbf{\#Backprop.} \\
     &  & \textbf{Per Iter} & \textbf{Per Iter} \\
    \midrule
    Algorithm~\ref{alg:singletask_supernet} & 1 & 1 & 1 \\
    \midrule
    Algorithm~\ref{alg:multitask_supernet} & $T$ & $T$ & $T$ \\
    \midrule
    Algorithm~\ref{alg:multitask_supernet_is} & $T$ & 1 & $T$ \\
    \midrule
    Algorithm~\ref{alg:multitask_supernet_is_reinforce} &  $T$ &  1 &  1 \\
    \bottomrule
    \end{tabular}
    }
  \label{tab:algs_comparision}
\end{table}

\subsubsection{Differentiable NAS for a Single Task}
We start from a typical differentiable NAS designed for a single task, which can be formulated as
\begin{equation}
\underset{\va \in \mathcal{A}, \vw}{\min} ~ \ell^t(\va, \vw), 
\label{eqn:nas}
\end{equation}
where $\va$ is a candidate architecture in the search space $\mathcal{A}$, $\vw$ is the supernet's weight, and $\ell^t(\cdot)$ is the loss function of task-$t$ that also considers the cost  of architecture $\va$.  Following \cite{wu2019fbnet,dai2019chamnet}, the cost of an architecture can be defined in terms of FLOPs, parameter size, latency, energy, \textit{etc.}

In our work, we search in a block-level search space. For block-$b$ of the supernet, we have
\begin{equation}
    \vy = a_b f_b(\vx) + (1 - a_b) \vx,
\end{equation}
where $\vx$, $\vy$ are input and output of block-$b$ function $f_b(\cdot)$. $a_b \in \{0, 1\}$ is a binary variable that determines whether to compute block-$b$ or skip it. Under this setting, the search space $\mathcal{A} = \{0, 1\}^B$ for Equation (\ref{eqn:nas}) is combinatorial and contains $2^B$ candidates, where $B$ is the number of blocks. To solve it efficiently, we relax the problem as 
\begin{equation}
    \underset{\vpi, \vw}\min ~ \mathbb{E}_{\va \sim p_{\vpi}} \{\ell^t(\va, \vw)\},
    \label{eqn:nas_relax}
\end{equation}
where $\va \in \{0, 1\}^B$ is a random variable sampled from a distribution $p_\vpi$, parameterized by $\vpi \in [0, 1]^B$. For each block, we independently sample $a_b \sim \text{Bernoulli}(\pi_b)$ from a Bernoulli distribution with an expected value of $\pi_b$. The probability of architecture $\va$ computes as
\begin{equation}
    p_{\vpi}(\va) = \prod_{b=1}^B \pi_b^{a_b}(1 - \pi_b)^{(1 - a_b)}.
    \label{eqn:prob}
\end{equation}

Under this relaxation, we can jointly optimize the supernet's weight $\vw$ and architecture parameter $\vpi$ with stochastic gradient descent. Specifically, in the forward pass, we first sample $\va \sim p_\vpi$, and compute the loss with input data $\vx$, weights $\vw$, and architecture $\va$. Next, we compute gradient with respect to $\vw$ and $\va$. Since architecture $\va$ is a discrete random variable, we cannot pass the gradient directly to $\vpi$. Previous works have adopted the Straight-Through Estimator\cite{bengio2013estimating} to approximate the gradient to $\vpi$ as 
$\frac{\partial l^t}{\partial \vpi} \approx \frac{\partial l^t}{\partial \va}$. 
Alternatively, Gumbel-Softmax \cite{jang2016categorical, maddison2016concrete,wu2019fbnet} can also be used to estimate the gradient. 
We train $\vw$ and $\vpi$ jointly using SGD with learning rate $\eta, \eta_\pi$. After the training finishes, we sample architectures $\va$ from the trained distribution $p_{\vpi}$ and pass them to target task's training pipeline. This process is summarized in Algorithm \ref{alg:singletask_supernet}. 

\begin{algorithm}
\caption{Differentiable NAS for a Single Task}
\label{alg:singletask_supernet}
\begin{algorithmic}[1]
  \FOR{iter = 1, $\cdots$, N}
  \STATE Sample a batch of data $\vx$
  \STATE Sample $\va \sim p_\vpi$
  \STATE Forward pass to compute $\ell^t(\va, \vw, \vx)$
  \STATE Backward pass to compute $\frac{\partial \ell^t}{\partial \vw}, \frac{\partial \ell^t}{\partial \va}$
  \STATE Straight-Through Estimation $\frac{\partial \ell^t}{\partial \vpi} \leftarrow \frac{\partial \ell^t}{\partial \va}$
  \STATE Gradient update $\vw \leftarrow \vw - \eta \frac{\partial \ell^t}{\partial \vw}, \vpi \leftarrow \vpi - \eta_\pi \frac{\partial \ell^t}{\partial \vpi}.$
  \ENDFOR
 \STATE Sample $\va \sim p_\vpi$ for target task
\end{algorithmic}
\end{algorithm}
\vspace{-1em}

\subsubsection{Extending to Multiple Tasks}
We are interested in searching architectures for multiple tasks, which can be formulated as 
\begin{equation}
    \underset{\va^1, \cdots, \va^T, \vw^1, \cdots \vw^T}{\min} ~ \sum_{t=1}^T \ell^t(\va^t, \vw^t).
\label{eqn:multitask_multirun}
\end{equation}
This is a rather awkward way to combine $T$ independent optimization problems together. To simplify the problem, we first approximate Equation (\ref{eqn:multitask_multirun}) as
\begin{equation}
\underset{\va^1, \cdots, \va^T, \vw}{\min} ~ \sum_{t=1}^T \ell^t(\va^t, \vw),  
\label{eqn:multitask_nas}
\end{equation}
where $\vw$ is the weight of an over-parameterized supernet shared among all tasks, and $\va^t$ is the architecture sampled for task-$t$. One concern of using Equation (\ref{eqn:multitask_nas}) to approximate Equation (\ref{eqn:multitask_multirun}) is that in multitask learning, the optimization of different tasks may interfere with each other. We conjecture that in an over-parameterized supernet with large enough capacity, the interference is small and can be ignored. Also, unlike conventional multitask learning, our goal is not to train a network with multitask capability, but to find optimal architectures $\va^t$ for each task. We conjecture that the task interference has limited impact on the search results.

Using the same relaxation trick as Equation (\ref{eqn:nas_relax}), we re-write Equation (\ref{eqn:multitask_nas}) as 
\begin{equation}
    \underset{\vpi^1, \cdots, \vpi^T, \vw}\min ~ \sum_{t=1}^T \mathbb{E}_{\va^t \sim p_{\vpi^t}} \{\ell^t(\va^t, \vw)\},
    \label{eqn:multitask_nas_relax}
\end{equation}
where $\va^t$ are architectures sampled from a task-specific distribution $p_{\vpi^t}$ parameterized by $\vpi^t$. 
To solve this, we can slightly modify Algorithm \ref{alg:singletask_supernet} to reach Algorithm \ref{alg:multitask_supernet}.
\begin{algorithm}
\caption{Differentiable NAS for multiple tasks}
\label{alg:multitask_supernet}
\begin{algorithmic}[1]
  \FOR{iter = 1, $\cdots$, N}
  \STATE Sample a batch of data $\vx$
  \FOR{task $t=1, \cdots, T$}
  \STATE Sample $\va^t \sim p_{\vpi^t}$
  \STATE Forward pass to compute $\ell^t(\va^t, \vw, \vx)$
  \STATE Backward pass to compute $\frac{\partial \ell^t}{\partial \vw}, \frac{\partial \ell^t}{\partial \va^t}$
  \STATE Accumulate $\Delta_{\vw} = \Delta_{\vw} + \frac{\partial \ell^t}{\partial \vw}$
  \STATE Straight-Through Estimation $\Delta_{\vpi^t}
  \leftarrow \frac{\partial \ell^t}{\partial \va^t}$
  \ENDFOR
  \STATE Gradient update $\vw \leftarrow \vw - \eta \Delta_{\vw}$
  \STATE Gradient update $\vpi^t \leftarrow \vpi^t - \eta_\pi \Delta_{\vpi^t}$ for $t=1, \cdots, T$
  \ENDFOR
 \STATE Sample $\va^t \sim p_{\vpi^t}$ and for target task-$t$
\end{algorithmic}
\end{algorithm}

With Algorithm \ref{alg:multitask_supernet}, we did not gain efficiency compared with running the Algorithm \ref{alg:singletask_supernet} for $T$ times, since we need to compute $T$ forward and backward passes in each iteration. With the same number of iterations, we end up with a $T$ times higher compute cost. But in the next two sections, we show how we adopt \textit{importance sampling} and \textit{REINFORCE} to reduce the number of forward and backward passes to $1$.

\subsubsection{Reducing $T$ Forward Passes to $1$}
Reviewing Algorithm \ref{alg:multitask_supernet}, the need to run multiple forward passes comes from lines 4 and 5 that for each task, we need to sample different architectures from different $p_{\vpi^t}$ to estimate the expected task loss $\mathbb{E}_{\va^t \sim p_{\vpi^t}} \{\ell^t(\va^t, \vw)\}$ under  $p_{\vpi^t}$. 

Using \textit{Importance Sampling} \cite{mcbook}, we reduce $T$ forward passes into $1$. Instead of sampling $T$ architectures from $T$ distributions, we can just sample architectures once from a common proxy distribution $q$ and let $T$ tasks share the same architecture $\va$ in their the forward pass. Though not sampling from $p_{\vpi^t}$, we can still compute an unbiased estimation of the task loss expectation $\mathbb{E}_{\va^t \sim p_{\vpi^t}} \{\ell^t(\va^t, \vw)\}$ as
\begin{equation}
\begin{gathered}
    \mathbb{E}_{\va^t \sim p_{\vpi^t}}\{\ell^t(\va^t, \vw)\}  = \mathbb{E}_{\va \sim q}\{\frac{p_{\vpi^t}(\va)}{q(\va)}\ell^t(\va, \vw)\} \\
     \approx \frac{1}{N}\sum_{i=1}^N \frac{p_{\vpi^t}(\va_i)}{q(\va_i)}\ell^t(\va_i, \vw), \text{ with } \va_i \sim q.
    \label{eqn:importance_sampling}
\end{gathered}
\end{equation}
$N$ is the number of architecture samples. $q$ can be any distribution as long as it satisfies the condition that $q(\va) \ne 0$ where $p_{\vpi^t}(\va) \ne 0$. Equation (\ref{eqn:importance_sampling}) will always be an unbiased estimator. We empirically design $q$ as a distribution that we first uniformly sample a task from $\{1, \cdots, T\}$, and sample the architecture from $p(\va)_{\vpi^t}$. For any architecture $\va$, its probability can be calculated as $q(\va) = 1 / T \sum_{t} p(\va)_{\vpi^t} $ with $ p(\va)_{\vpi^t}$ computed by Equation (\ref{eqn:prob}). Using importance sampling, we redesign the search algorithm as Algorithm \ref{alg:multitask_supernet_is} to reduce the number of forward passes from $T$ to 1. 

\begin{algorithm}
\caption{Reduce forward passes with Importance Sampling}
\label{alg:multitask_supernet_is}
\begin{algorithmic}[1]
  \FOR{iter = 1, $\cdots$, N}
  \STATE Sample a batch of data $\vx$
  \STATE Sample $\va \sim q$
  \STATE Forward pass to compute $\vy = f(\va, \vw, \vx)$
  \FOR{task $t=1, \cdots, T$}
  \STATE Importance Sampling $\gamma_t \leftarrow p_{\vpi^t}(\va) / q(\va)$
  \STATE Compute task loss $\ell^t \leftarrow \gamma_t \times \ell^t(\va, \vw, \vy)$
  \STATE Backward pass to compute $\frac{\partial \ell^t}{\partial \vw}, \frac{\partial \ell^t}{\partial \va}$
  \STATE Accumulate $\Delta_{\vw} = \Delta_{\vw} + \frac{\partial \ell^t}{\partial \vw}$
  \STATE Straight-Through Estimation $\Delta_{\vpi^t}
  \leftarrow \frac{\partial \ell^t}{\partial \va}$
  \ENDFOR
  \STATE Gradient update $\vw \leftarrow \vw - \eta \Delta_{\vw}$ 
  \STATE Gradient update $\vpi^t \leftarrow \vpi^t - \eta_\pi \Delta_{\vpi^t}$ for $t=1, \cdots, T$
  \ENDFOR
 \STATE Sample $\va^t \sim p_{\vpi^t}$ for target task
\end{algorithmic}
\end{algorithm}
\vspace{-1em}

\subsubsection{Reducing $T$ Backward Passes to $1$}
Algorithm \ref{alg:multitask_supernet_is} only requires 1 forward pass but $T$ backward passes. This ie because to optimize the architecture distribution for task-$t$, we need to run a backward pass to compute $\partial \ell^t / \partial \va$, which we use to estimate $\partial \ell^t / \partial \vpi ^t$ and to update the task architecture parameter $\vpi^t$. To avoid this, we use REINFORCE \cite{williams1992simple} to estimate the gradient  $\partial \ell^t / \partial \vpi ^t$ as
\begin{equation}
\begin{gathered}
    \nabla_{\vpi^t} \mathbb{E}_{\va \sim p_{\vpi^t}} \{\ell^t(\va)\}  = \nabla_{\vpi^t} \sum_{\va \in \mathcal{A}} p_{\vpi^t}(\va) \ell^t(\va) \\
    = \sum_{\va \in \mathcal{A}} \ell^t(\va) \nabla_{\vpi^t}  p_{\vpi^t}(\va) 
 =  \sum_{\va \in \mathcal{A}} \ell^t(\va)   p_{\vpi^t}(\va) \nabla_{\vpi^t} \log p_{\vpi^t}(\va) \\
 = 
 \mathbb{E}_{\va \sim p_{\vpi^t}} \{\ell^t(\va)  \nabla_{\vpi^t} \log p_{\vpi^t}(\va) \} \\
 \approx \frac{1}{N} \sum_{i=1}^N \ell^t(\va_i) \nabla_{\vpi^t} \log p_{\vpi^t}(\va_i), \text{ with } \va_i \sim p_{\vpi^t}.
 \label{eqn:reinforce}
\end{gathered}
\end{equation}
$N$ is the number of architecture samples. Given the definition of $p_{\vpi^t}(\va)$ in Equation (\ref{eqn:prob}), we can easily derive $\nabla_{\vpi^t} \log p_{\vpi^t}(\va)$, with its $b$-th element simply computed as 
\begin{equation}
    (\nabla_{\vpi^t} \log p_{\vpi^t}(\va))_b = 1 / (\pi_b^{a_b}(1 - \pi_b)^{(1 - a_b)})).
\end{equation}
 Equation (\ref{eqn:reinforce}) is also referred to as the \textit{score function estimator} of the true gradient $\partial \ell^t / \partial \vpi ^t$. 
The intuition is that for any sampled architecture $\va_i$, we score its gradient by the loss $\ell^t(\va_i)$, such that architectures that cause larger loss will be suppressed and vice versa. 
This technique is more often referred to as the \textit{policy gradient} in Reinforcement Learning. For NAS, a similar technique is adopted by \cite{casale2019probabilistic,yan2021fp} to search for classification models. Using Equation (\ref{eqn:reinforce}), we no longer need to run back propogation to compute $\partial \ell^t / \partial \vpi ^t$ for each task. We still need to compute the gradient to the supernet weights $\partial \ell / \partial \vw$, but we can first sum up the task losses $\ell^t$ and run backward pass only once. This is summarized in Algorithm \ref{alg:multitask_supernet_is_reinforce} and visualized in Figure \ref{fig:overview}. We discuss more important details of this algorithm in Appendix~\ref{sec:alg_details}. Note that we still have two for-loops in each iteration to compute the task loss $\gamma_t \ell^t$ from the network's prediction $\vy$ and the gradient estimator for $\partial \ell^t /\partial \vpi ^t$, but their computational cost is negligible compared with the forward and backward passes. 
 
\begin{algorithm}
\caption{A Single Run Multitask NAS with Importance Sampling and REINFORCE}
\label{alg:multitask_supernet_is_reinforce}
\begin{algorithmic}[1]
  \FOR{iter = 1, $\cdots$, N}
  \STATE Sample a batch of data $\vx$
  \STATE Sample $\va \sim q$
  \STATE Forward pass $\vy = f(\va, \vw, \vx)$
  \FOR{task $t=1, \cdots, T$}
  \STATE Importance Sampling $\gamma_t \leftarrow p_{\vpi^t}(\va) / q(\va)$
  \STATE Accumulate loss $\ell = \ell +\gamma_t \times \ell^t(\va, \vw, \vy) $
  \ENDFOR
  \STATE Backward pass to compute $\vw \leftarrow \vw - \eta \frac{\partial \ell}{\partial \vw}$
  \FOR{task $t=1, \cdots, T$}
  \STATE
  REINFORCE $\vpi^t \leftarrow \vpi^t - \eta_{\pi} \ell^t \nabla_{\vpi^t}\log p_{\vpi^t}(\va) $
  \ENDFOR
  \ENDFOR
 \STATE Sample $\va^t \sim p_{\vpi^t}$ for target task
\end{algorithmic}
\end{algorithm}

\vspace{-0.5em}
\section{Experiments}

\subsection{Experiment Settings}
\label{sec:exp_settings}
We implement the search process and target task's training pipeline in D2Go\footnote{https://github.com/facebookresearch/d2go} powered by Pytorch~\cite{paszke2019pytorch} and Detectron2~\cite{wu2019detectron2}. For the \underline{search (training supernet) process}, we build a supernet extended from an FBNetV3-A model as illustrated in Section~\ref{sec:space}. During search, we first pretrain the supernet on ImageNet~\cite{deng2009imagenet} with classification labels for 1100 epochs, mostly following a regular classification training recipe~\cite{graham2021levit,touvron2020training}. More details are included in Appendix~\ref{sec:train_details}. This step takes about 60 hours to finish on 64 V100 GPUs.
Then we train the supernet on the multitask proxy dataset for 9375 steps using SGD with a base learning rate of 0.96. We decay the learning rate by 10x at step-3125. We set the initial sampling probability of all blocks to 0.5. We do not update the architecture parameters until step-6250. We set the architecture parameter's learning rate to be 0.01 of the weight's learning rate. It takes about 10 hours to finish when trained on 16 V100 GPUs. More details of the search implementation can be found in Appendix~\ref{sec:search_details}. After the search, we sample the most likely architectures for each task.

For \underline{training the searched architectures}, we mostly follow existing SotA training recipes for each task~\cite{graham2021levit,cheng2021maskformer,wu2019detectron2}. See Appendix~\ref{sec:train_details} for details. For semantic segmentation, we follow MaskFormer~\cite{cheng2021maskformer} and attach a modified light-weight MaskFormer head (dubbed Lite MaskFormer) to the searched backbone. For object detection, we use Faster R-CNN's~\cite{ren2015faster} detection head with light-weight ROI and RPN. We call the new head as Lite R-CNN. See the architecture design of the two light-weight heads in Appendix \ref{sec:head_arch}. 

\subsection{Comparing with SotA Compact Models}

We compare our searched architectures against both NAS searched and manually designed compact models for ImageNet  \cite{deng2009imagenet} classification, ADE20K~\cite{zhou2017scene} semantic segmentation, and COCO~\cite{lin2014microsoft} object detection.
We search topologies for all tasks by training supernet once, sampling one topology for each task, and transfer the searched topology to different versions of FBNetV3 models with different sizes. We use FBNetV3-\{A, C, F\} and build two smaller models FBNetV3-A$_R$ and FBNetV3-A$_C$ by mainly shrinking the resolution and channel sizes from FBNetV3-A, respectively.
See Appendix \ref{sec:shrunken_fbnetv3_a}. We name a model using the template {\FrameworkName}-\texttt{\{version\}}-\texttt{\{task\}}. For a given task, all models share the same searched topology, as in Figure \ref{fig:vis_search}.

Compared with all the existing compact models including automatically searched and manually designed ones, our {\FrameworkName} delivers architectures with better accuracy/mIoU/mAP vs. efficiency trade-offs in all the ImageNet~\cite{deng2009imagenet} classification (\eg, $\uparrow$1.3\% top-1 accuracy under the same FLOPs as compared to FBNetV3-G~\cite{wu2019fbnet}), ADE20K~\cite{zhou2017scene} segmentation (\eg, $\uparrow$1.8\% higher mIoU than SegFormer  with MiT-B1 as backbone~\cite{xie2021segformer} and 3.6$\times$ fewer FLOPs), and COCO~\cite{lin2014microsoft} detection tasks (\eg, $\uparrow$1.1\% mAP with 1.2$\times$ fewer FLOPs as compared to YOLOX-Nano~\cite{nanodet}). See Tables \ref{tab:imagenet}, \ref{tab:ade20k}, \ref{tab:coco} and Figure \ref{fig:performance} for a detailed comparison. 

\begin{table}[t]
\caption{Comparisons with SotA compact models on the \textbf{ImageNet}~\cite{deng2009imagenet} image classification task.}
\centering
  \resizebox{0.95\linewidth}{!}
  {
    \begin{tabular}{cc||cc}
    \toprule
     \multirow{2}{*}{\textbf{Model}} & \textbf{Input} &  \multirow{2}{*}{\textbf{FLOPs}} &\textbf{Accuracy} \\
      & \textbf{Size}   &  &\textbf{(\%, Top-1)} \\
     \midrule
     HR-NAS-A~\cite{ding2021hr} & 224 $\times$ 224 & 267M & 76.6 \\
     LeViT-128S~\cite{graham2021levit} & 224 $\times$ 224 & 305M & 76.6 \\
     BigNASModel-S~\cite{yu2020bignas}& 192 $\times$ 192 & 242M & 76.5 \\
     MobileNetV3-1.25x~\cite{howard2019searching} & 224 $\times$ 224 & 356M & 76.6 \\
    \textbf{{\FrameworkName}-A$_{R}$-CLS}  & 160 $\times$ 160 & \textbf{215M} & \textbf{77.2} \\
     \midrule
     HR-NAS-B~\cite{ding2021hr} & 224 $\times$ 224 & 325M & 77.3 \\
     LeViT-128~\cite{graham2021levit} & 224 $\times$ 224 & 406M & 78.6 \\
     EfficientNet-B0~\cite{tan2019efficientnet}  & 224 $\times$ 224 & 390M & 77.3 \\
     \textbf{{\FrameworkName}-A$_{C}$-CLS} & 224 $\times$ 224 & \textbf{280M} & \textbf{78.4} \\
     \midrule
      EfficientNet-B1~\cite{tan2019efficientnet} & 240 $\times$ 240 & 700M & 79.1 \\
      FBNetV3-E~\cite{Dai2020FBNetV3JA} & 264 $\times$ 264 & 762M & 81.3 \\
      \textbf{{\FrameworkName}-A-CLS} & 224 $\times$ 224 & \textbf{685M} & \textbf{81.7} \\
     \midrule
     LeViT-256~\cite{graham2021levit} & 224 $\times$ 224 & 1.1G & 81.6 \\
     EfficientNet-B2~\cite{tan2019efficientnet} & 260 $\times$ 260 & 1.0G & 80.3 \\
     BigNASModel-XL~\cite{yu2020bignas} & 288 $\times$ 288 & 1.0G & 80.9 \\
     FBNetV3-F~\cite{Dai2020FBNetV3JA} & 272 $\times$ 272 &  1.2G & 82.5 \\
     \textbf{{\FrameworkName}-C-CLS} & 248 $\times$ 248 & \textbf{1.0G} & \textbf{82.6} \\
     \midrule
     Swin-T~\cite{liu2021swin} & 224 $\times$ 224 & 4.5G & 81.3 \\
     LeViT-384~\cite{graham2021levit} & 224 $\times$ 224 & 2.4G & 82.6 \\
     BossNet-T1~\cite{li2021bossnas}  & 288 $\times$ 288 & 5.7G & 81.6 \\
     EfficientNet-B4~\cite{tan2019efficientnet} & 380 $\times$ 380 & 4.2G & 82.9 \\
     FBNetV3-G~\cite{Dai2020FBNetV3JA} & 320 $\times$ 320   & 2.1G & 82.8 \\
     \textbf{{\FrameworkName}-F-CLS} & 272 $\times$ 272 & \textbf{2.1G} & \textbf{84.1} \\
    \bottomrule
    \end{tabular}
    }
  \label{tab:imagenet}
\end{table}

\begin{table}[t]
\caption{Comparisons with SotA compact models on the \textbf{ADE20K} semantic segmentation task. All mIoUs are reported in the single-scale setting (except those marked with $^{\dagger}$) in ADE20K \textit{val.} and FLOPs is measured with the input resolution of \texttt{(short\_size $\times$ short\_size)} following~\cite{cheng2021maskformer,xie2021segformer}. The implementation details of Lite MaskFormer are illustrated in Appendix \ref{sec:head_arch}.}
  \resizebox{1.0\linewidth}{!}
  {
    \begin{tabular}{ccc||ccc}
    \toprule
     \multirow{2}{*}{\textbf{Backbone}} & \multirow{2}{*}{\textbf{Head}} & \textbf{Short} & \multirow{2}{*}{\textbf{FLOPs}} &\textbf{mIoU} \\
       & & \textbf{Size} & & \textbf{(\%)} \\
      \midrule
     HR-NAS-A~\cite{ding2021hr} & Concatenation~\cite{ding2021hr} & 512& 1.4G & 33.2 \\
     MobileNetV3-Large~\cite{li2021searching} & Lite MaskFormer & 448 & 1.5G & 29.2 \\
     \textbf{{\FrameworkName}-A$_C$-SEG} & Lite MaskFormer  & 384 & \textbf{1.3G} & \textbf{35.6} \\
     \midrule
     HR-NAS-B~\cite{ding2021hr} & Concatenation~\cite{ding2021hr} & 512 & 2.2G & 34.9 \\
    EfficientNet-B0~\cite{tan2019efficientnet} & Lite MaskFormer & 448 & 2.1G & 31.3 \\
    \textbf{{\FrameworkName}-A$_R$-SEG} & Lite MaskFormer  & 384 & \textbf{1.8G} & \textbf{37.8} \\
     \midrule
     MiT-B0~\cite{xie2021segformer} & SegFormer~\cite{xie2021segformer} & 512 & 8.4G & 37.4 \\
    \textbf{{\FrameworkName}-A-SEG} & Lite MaskFormer & 384 & \textbf{2.9G} & \textbf{41.2} \\
     \midrule
     MiT-B1~\cite{xie2021segformer} & SegFormer~\cite{xie2021segformer} & 512& 15.9G & 42.2 \\
     \textbf{{\FrameworkName}-C-SEG} & Lite MaskFormer & 448 & \textbf{4.4G} & \textbf{44.0} \\
     \midrule
     Swin-T~\cite{liu2021swin} & UperNet~\cite{xiao2018unified} & 512 & 236G & 46.1$^{\dagger}$ \\
     Swin-T~\cite{liu2021swin} & MaskFormer~\cite{cheng2021maskformer}& 512  & 55G & 46.7 \\
     ResNet-50~\cite{he2016deep} & MaskFormer~\cite{cheng2021maskformer} & 512 & 53G & 44.5 \\
     PVT-Large~\cite{wang2021pyramid} & Semantic FPN~\cite{kirillov2019panoptic} & 512 & 80G & 44.8$^{\dagger}$ \\
    \textbf{{\FrameworkName}-F-SEG} & Lite MaskFormer & 512 & \textbf{9.4G} & \textbf{46.5} \\
    \bottomrule
    \vspace{-2em}
    \end{tabular}
    }
  \label{tab:ade20k}
\end{table}

\begin{table}[t]
\caption{Comparisons with SotA compact models on the \textbf{COCO} object detection task. mAPs are based on COCO \textit{val}. For \cite{ge2021yolox,nanodet,xiong2020mobiledets}, we cite their FLOPs with the given resolution. For R-CNN models, since their input sizes are not fixed, we report the \textit{average} FLOPs on the COCO \textit{val} dataset. See Appendix \ref{sec:avg_flop} for details.}

\centering
  \resizebox{1.0\linewidth}{!}
  {
    \begin{tabular}{ccc||cc}
    \toprule
     \multirow{2}{*}{\textbf{Backbone}} & \multirow{2}{*}{\textbf{Head}}  &
     \textbf{Short, Long} & \multirow{2}{*}{\textbf{FLOPs}} &\textbf{mAP} \\
       &  & \textbf{Size} & &\textbf{(\%)} \\
     \midrule
     ShuffleNetV2 1.0x~\cite{ma2018shufflenet} & NanoDet-m~\cite{nanodet} & 320, 320  & 720M & 20.6 \\
     EfficientNet-B0~\cite{tan2019efficientnet} & Lite R-CNN & 224, 320 & 793M & 23.1 \\
     \textbf{{\FrameworkName}-A$_C$-DET} & Lite R-CNN & 224, 320 & \textbf{713M} & \textbf{25.0} \\
     \midrule
     MobileDets~\cite{xiong2020mobiledets} & SSDLite~\cite{sandler2018mobilenetv2} & 320, 320  & 920M & 25.6 \\
     ShuffleNetV2 1.0x~\cite{ma2018shufflenet} & NanoDet-m~\cite{nanodet} & 416, 416 & 1.2G & 23.5 \\
     Modified CSP v5~\cite{ge2021yolox} & YOLOX-Nano~\cite{ge2021yolox} & 416, 416  & 1.1G & 25.3 \\
     EfficientNet-B2~\cite{tan2019efficientnet} & Lite R-CNN & 224, 320 & 1.2G & 24.9 \\
     \textbf{{\FrameworkName}-A$_R$-DET} & Lite R-CNN & 224, 320 & \textbf{908M} & \textbf{26.4} \\
     \midrule
     ShuffleNetV2 1.5x~\cite{ma2018shufflenet} & NanoDet-m~\cite{nanodet} & 416, 416 & 2.4G & 26.8 \\
     EfficientNet-B3~\cite{tan2019efficientnet} & Lite R-CNN & 224, 320 & 1.6G & 26.2 \\
     \textbf{{\FrameworkName}-A-DET} & Lite R-CNN & 224, 320  & \textbf{1.35G} & \textbf{27.2} \\
     \textbf{{\FrameworkName}-A$_C$-DET} & Lite R-CNN & 320, 640 & \textbf{1.37G} & \textbf{28.9} \\
     \textbf{{\FrameworkName}-A$_R$-DET} & Lite R-CNN & 320, 640 & \textbf{1.80G} & \textbf{30.4} \\
    \bottomrule
    \vspace{-2em}
    \end{tabular}
    }
  \label{tab:coco}
\end{table}

\begin{table}[b]
\vspace{-0.5em}
\caption{Effectiveness of our search algorithms when benchmarked in ImageNet~\cite{deng2009imagenet} image classification (\texttt{CLS}), ADE20K~\cite{zhou2017scene} semantic segmentation (\texttt{SEG}), and COCO~\cite{lin2014microsoft} object detection (\texttt{DET}). $T$ represents the number of tasks. All the models of \texttt{SEG} are trained for 160K iterations for fast verification.}
\centering
  \resizebox{\linewidth}{!}
  {
    \begin{tabular}{c||cc||cc}
    \toprule
     \multirow{2}{*}{\textbf{Tasks}} & \textbf{Search} &  \textbf{Search Cost} & \multirow{2}{*}{\textbf{FLOPs}} &\textbf{Top-1 Accuracy/} \\
      & \textbf{Algorithm} &  \textbf{(GPU hours)} &  &\textbf{mIoU / mAP (\%)} \\
     \midrule
     \multirow{3}{*}{\texttt{CLS}} & Random & - & 769M & 81.5 \\
     &  Single Task (Alg.~\ref{alg:singletask_supernet})  & 4000 & 688M & 81.9 ($\uparrow$0.4) \\
     &  \textbf{{\FrameworkName} (Alg.~\ref{alg:multitask_supernet_is_reinforce})}  & \textbf{4000 / $T$} & \textbf{726M} & \textbf{81.8 ($\uparrow$0.3)} \\
    \midrule
     \multirow{3}{*}{\texttt{SEG}} & Random & - & 2.9G & 38.8 \\
     &   Single Task (Alg.~\ref{alg:singletask_supernet}) & 4000 & 2.7G & 40.4 ($\uparrow$1.6) \\
      &  \textbf{{\FrameworkName} (Alg.~\ref{alg:multitask_supernet_is_reinforce})}  & \textbf{4000 / $T$} & \textbf{2.8G} & \textbf{40.4 ($\uparrow$1.6)}\\
    \midrule
     \multirow{3}{*}{\texttt{DET}} & Random & - & 1.34G & 26.8 \\
     &   Single Task (Alg.~\ref{alg:singletask_supernet})  & 4000 & 1.36G & 27.3 ($\uparrow$0.5) \\
      & \textbf{{\FrameworkName} (Alg.~\ref{alg:multitask_supernet_is_reinforce})}  & \textbf{4000 / $T$} & \textbf{1.36G} & \textbf{27.2 ($\uparrow$0.4)}\\

    \bottomrule
    \end{tabular}
    }
  \label{tab:compare_single_random}
\end{table}

\subsection{Ablation Study on {\FrameworkName}'s Search Algorithm}

To verify the effectiveness of the search algorithm proposed in Section~\ref{sec:search-algorithm} (\ie, Algorithm~\ref{alg:multitask_supernet_is_reinforce}), we compare the proposed multitask search (Algorithm~\ref{alg:multitask_supernet_is_reinforce}) with single-task search (Algorithm~\ref{alg:singletask_supernet}) and random search. We sample four architectures from two trained distributions (by Algorithm~\ref{alg:multitask_supernet_is_reinforce} and Algorithm~\ref{alg:singletask_supernet}) and a random distribution where each block has a 0.5 probability being sampled. We compare sampled architectures with their best accuracy/mIoU/mAP vs. efficiency trade-off and report the results in Table~\ref{tab:compare_single_random}. First, random architectures achieve strong performance. This demonstrates the effectiveness of the search space design. But compared to the random search, using the same FLOPs, models from multitask search obviously outperforms randomly sampled models by achieving $\uparrow$0.3\% higher accuracy on image classification, $\uparrow$1.6\% higher mIoU on semantic segmemtation, and $\uparrow$0.4\% higher mAP on object detection.  Compared with single-task search, models searched by multitask search deliver very similar performance (\eg, 2.8G vs. 2.7G FLOPs under the same mIoU on ADE20K~\cite{zhou2017scene}) while reducing the search cost for each task by a factor of $T$ times.

\subsection{Searched Architectures for Different Tasks}

\begin{figure}[t]
  \centering
  \includegraphics[width=1.0\linewidth]{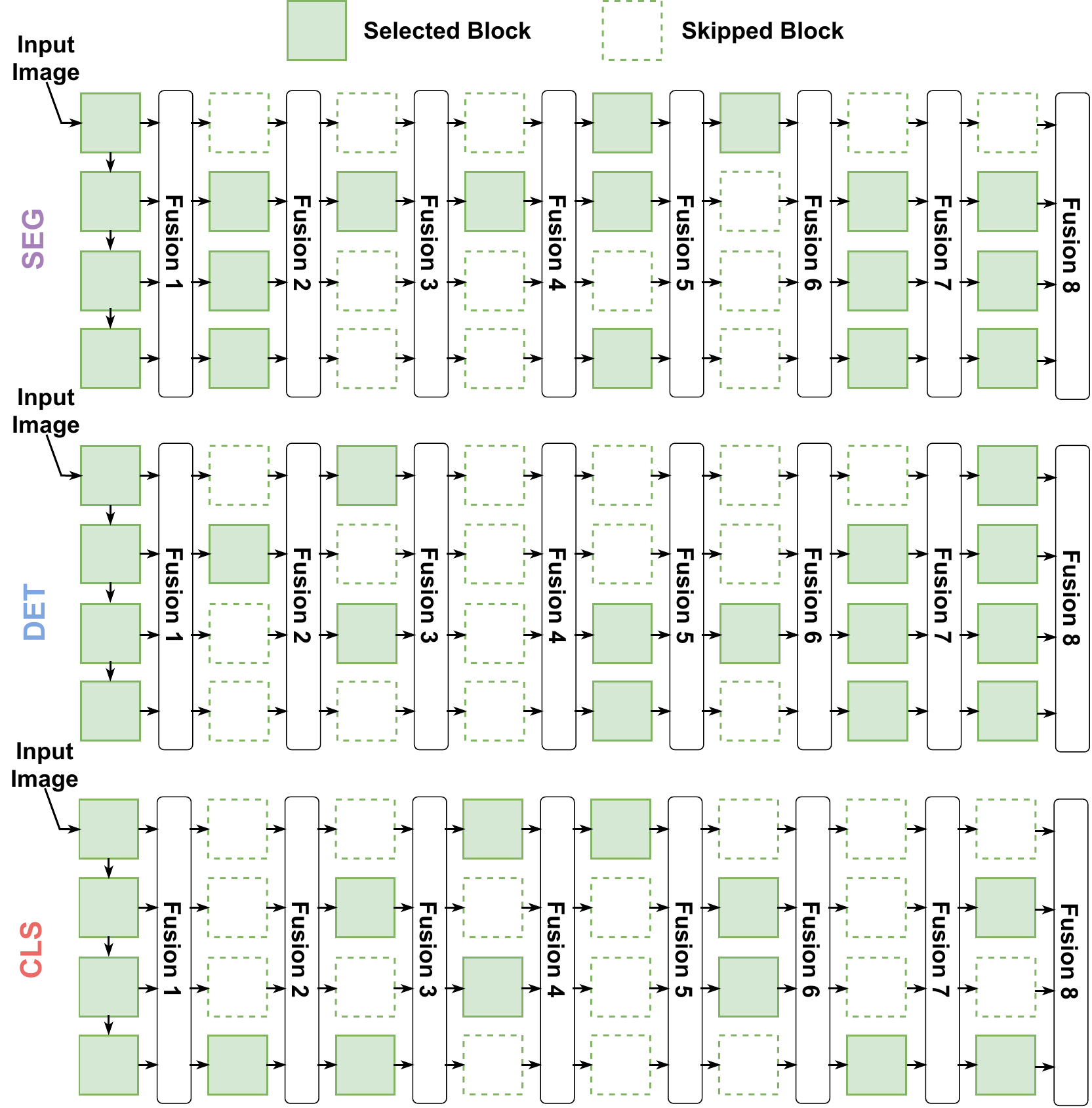}
\vspace{-1.5em}
\caption{Visualization of the searched architectures for semantic segmentation (\texttt{SEG}), object detection (\texttt{DET}), and image classification (\texttt{CLS}) tasks.}
\vspace{-1.em}
\label{fig:vis_search}
\end{figure}

To better understand the architectures searched by {\FrameworkName}, we visualize them in Figure~\ref{fig:vis_search}. For the \underline{\texttt{SEG}} model (Figure \ref{fig:vis_search}-top), its blocks between \texttt{Fusion 1} and \texttt{Fusion 6} match the U-Net's pattern that gradually increases feature resolutions. See Figure \ref{fig:special_cases}-top for a comparison. For the \underline{\texttt{DET}} model (Figure \ref{fig:vis_search}-middle), we did not find an obvious pattern to describe it. We leave the interpretation to each reader. 
Surprisingly, the \underline{\texttt{CLS}} model contains a lot of blocks from higher resolutions. This contrasts the mainstream models~\cite{wu2019fbnet,cai2018proxylessnas,cai2019once,li2021searching,yu2020bignas} that only stack layers sequentially. Given that our searched \texttt{CLS} model demonstrates stronger performance than sequential architectures, this may open up a new direction for the classification model design.

\section{Conclusion}
We propose {\FrameworkName}, a NAS framework that can search for neural architectures for a variety of CV tasks with reduced human effort and compute cost. {\FrameworkName} features a simple yet inclusive and transferable search space, a multi-task search process disentangled with target tasks' training pipelines, and a novel search algorithm with a constant compute cost agnostic to number of tasks. Our experiments show that in a single run of search,  {\FrameworkName} produces efficient models that significantly outperform the previous SotA models in ImageNet classification, COCO object detection, and ADE20K semantic segmentation.
\section{Discussion on Limitations}
There are several limitations of our work. First, we did not explore a more granular search space, \textit{e.g.}, to search for block-wise channel sizes, which can further improve searched models' performance. Second, while our framework can search for multiple tasks in one run, we do not support adding new tasks incrementally, which will further improve the task-scalability. One potential solution is to explore whether we can transfer the searched architectures  from one task (\eg, segmentation) to similar tasks (\eg, depth estimation) without re-running the search.
\newpage
{\small
\bibliographystyle{ieee_fullname}
\bibliography{egbib}
}
\clearpage
\clearpage
\appendix

\section{Block Configurations of FBNetV3-A Supernet}
\label{sec:block_config}
To explain how to extend an FBNetV3~\cite{Dai2020FBNetV3JA} to the supernet in \FrameworkName, we list the code snippets below. It includes the block configurations of both FBNetV3-A and the supernet extended from FBNetV3-A. It is compatible with with official implementation of FBNetV3~\footnote{https://github.com/facebookresearch/mobile-vision/blob/main/mobile\_cv/arch/fbnet\_v2/fbnet\_modeldef\_cls\_fbnetv3.py}. 
\begin{lstlisting}[language=Python, caption=Code snippets of FBNetV3-A]
# Official FBNetV3-A
# input_size: 224
# [[Operator, Channels, Stride, Repeats]]
# Partition 0
[["conv_k3_hs",16,2,1]],
[["ir_k3_hs",16,1,2,{"expansion": 1}]],
[
  ["ir_k5_hs",24,2,1,{"expansion": 4}],
  ["ir_k5_hs",24,1,3,{"expansion": 2}],
],
# Partition 1
[
  ["ir_k5_sehsig_hs",40,2,1,{"expansion": 5}],
  ["ir_k5_sehsig_hs",40,1,4,{"expansion": 3}],
],
# Partition 2
[
  ["ir_k5_hs",72,2,1,{"expansion": 5}],
  ["ir_k3_hs",72,1,4,{"expansion": 3}],
  ["ir_k3_sehsig_hs",120,1,1,{"expansion": 5}],
  ["ir_k5_sehsig_hs",120,1,5,{"expansion": 3}],
],
# Partition 3
[
  ["ir_k3_sehsig_hs",184,2,1,{"expansion": 6}],
  ["ir_k5_sehsig_hs",184,1,5,{"expansion": 4}],
  ["ir_k5_sehsig_hs",224,1,1,{"expansion": 6}],
],
\end{lstlisting}

\begin{lstlisting}[language=Python, caption=Code snippets of the supernet extended from FBNetV3-A]
# Supernet extended from FBNetV3-A
# input_size: 224,
# [[Operator, Channels, Stride, Repeats]]
# Path 0, Stage 0
[["conv_k3_hs",16,2,1]],
[["ir_k3_hs",16,1,2,{"expansion": 1}]],
[
  ["ir_k5_hs",24,2,1,{"expansion": 4}],
  ["ir_k5_hs",24,1,3,{"expansion": 2}],
],
# Path 1, Stage 0
[
  ["ir_k5_sehsig_hs",40,2,1,{"expansion": 5}],
  ["ir_k5_sehsig_hs",40,1,4,{"expansion": 3}],
],
# Path 2, Stage 0
[
  ["ir_k5_hs",72,2,1,{"expansion": 5}],
  ["ir_k3_hs",72,1,4,{"expansion": 3}],
  ["ir_k3_sehsig_hs",120,1,1,{"expansion": 5}],
  ["ir_k5_sehsig_hs",120,1,5,{"expansion": 3}],
],
# Path 3, Stage 0
[
  ["ir_k3_sehsig_hs",184,2,1,{"expansion": 6}],
  ["ir_k5_sehsig_hs",184,1,5,{"expansion": 4}],
  ["ir_k5_sehsig_hs",224,1,1,{"expansion": 6}],
],
# Path 0, Stage 1 to Stage s
[["ir_k5_hs",24,1,2,{"expansion": 2}]],
# Path1, Stage 1 to Stage s
[["ir_k5_sehsig_hs",40,1,2,{"expansion": 3}]],
# Path2, Stage 1 to Stage s
[["ir_k5_sehsig_hs",120,1,2,{"expansion": 3}]],
# Path3, Stage 1 to Stage s
[
  ["ir_k5_sehsig_hs",224,1,1,{"expansion": 4}],
  ["ir_k5_sehsig_hs",224,1,1,{"expansion": 6}],
],
# Fusion
[["fusion", [24, 40, 120, 224], 1, 1]],
\end{lstlisting}

\section{Details about the Fusion Module}

Following HRNet~\cite{wang2020deep}, in the supernet and searched network, we design the \texttt{Fusion} module to fuse feature maps with different resolutions with each other. The original HRNet's fusion modules are computationally expensive. To reduce cost, we design a parameter-free and (almost) compute-free fusion module. 

\begin{figure}[h]
  \centering
  \includegraphics[width=1.0\linewidth]{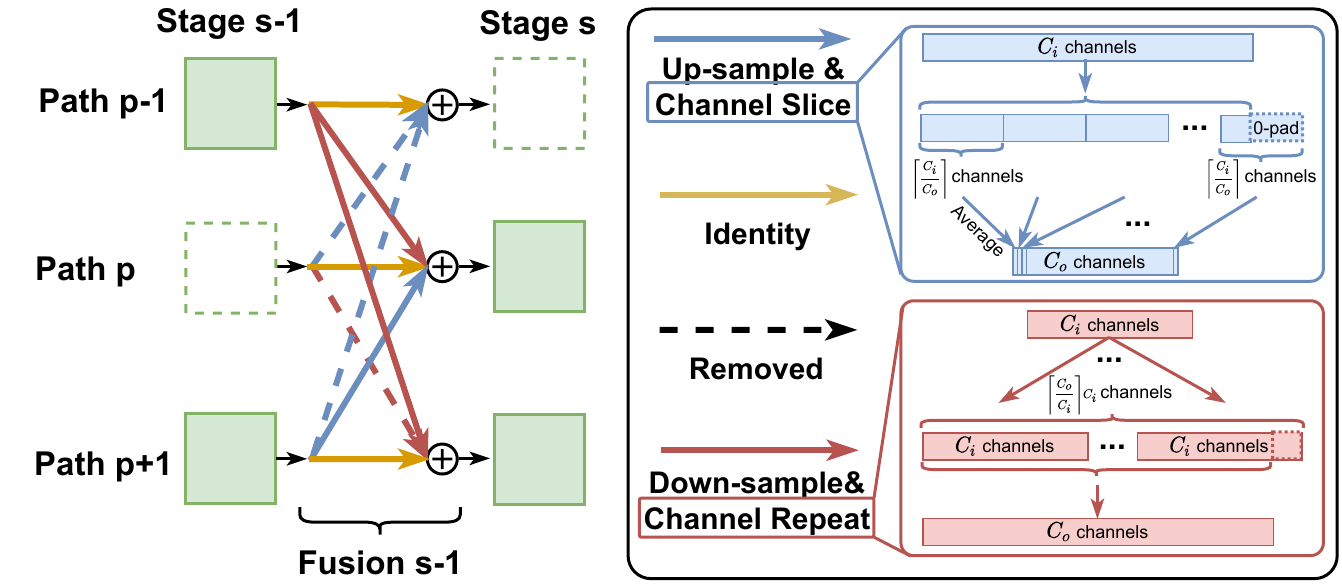}
\vspace{-2em}
\caption{Illustration of the fusion module aggregating information from different paths (resolutions).}
\label{fig:fusion}
\end{figure}

Each block in the network is fused to all blocks at the next stage. For a block at a given path: 1) if the output feature is at the same path (resolution), the fusion module is essentially a identity connection (blue arrows in Figure \ref{fig:fusion}). 2) To fuse the feature to a path with larger channel size and lower resolution, we first down-sample the input feature to the target size, and repeat the original channels by $ \left\lceil C_{out} / C_{in} \right\rceil $ times, where $C_{in}, C_{out}$ are input/output channel sizes. If $C_{out}$ is not divisible by  $C_{in}$, we drop the extra channels. This is shown as the red arrows in Figure \ref{fig:fusion}. 3) To fuse to a feature with higher resolution and smaller channel sizes, we first up-sample the feature map. Then, we pad the input feature's channels with zero such that the channel size becomes $C'_{in} =  \left\lceil C_{in} / C_{out} \right\rceil \times C_{out} $. Finally, we take every $C'_{in} / C_{out}$ channels as a group and compute a channel-wise average to produce a new output channel. This is shown as Figure \ref{fig:fusion} blue arrows. Features fused to the same block will be summed together as input to the block. 

This fusion module does not require any parameters, and only requires a negligible amount of compute for down-sampling, up-sampling, padding, and channel-wise average.

In the supernet and the searched architectures, if any block (\eg, the ones in \texttt{(Stage s-1, Path p)} or \texttt{(Stage s, Path p-1)}) is skipped, the corresponding connections in the fusion modules from and to the block will also be removed, except the connections from and to other blocks in the same path. 
\label{sec:fusion_details}

\section{Visualization of the Mainstream Topologies for CV Tasks}
\label{sec:special_cases}

\begin{figure}[t]
  \centering
  \includegraphics[width=1.0\linewidth]{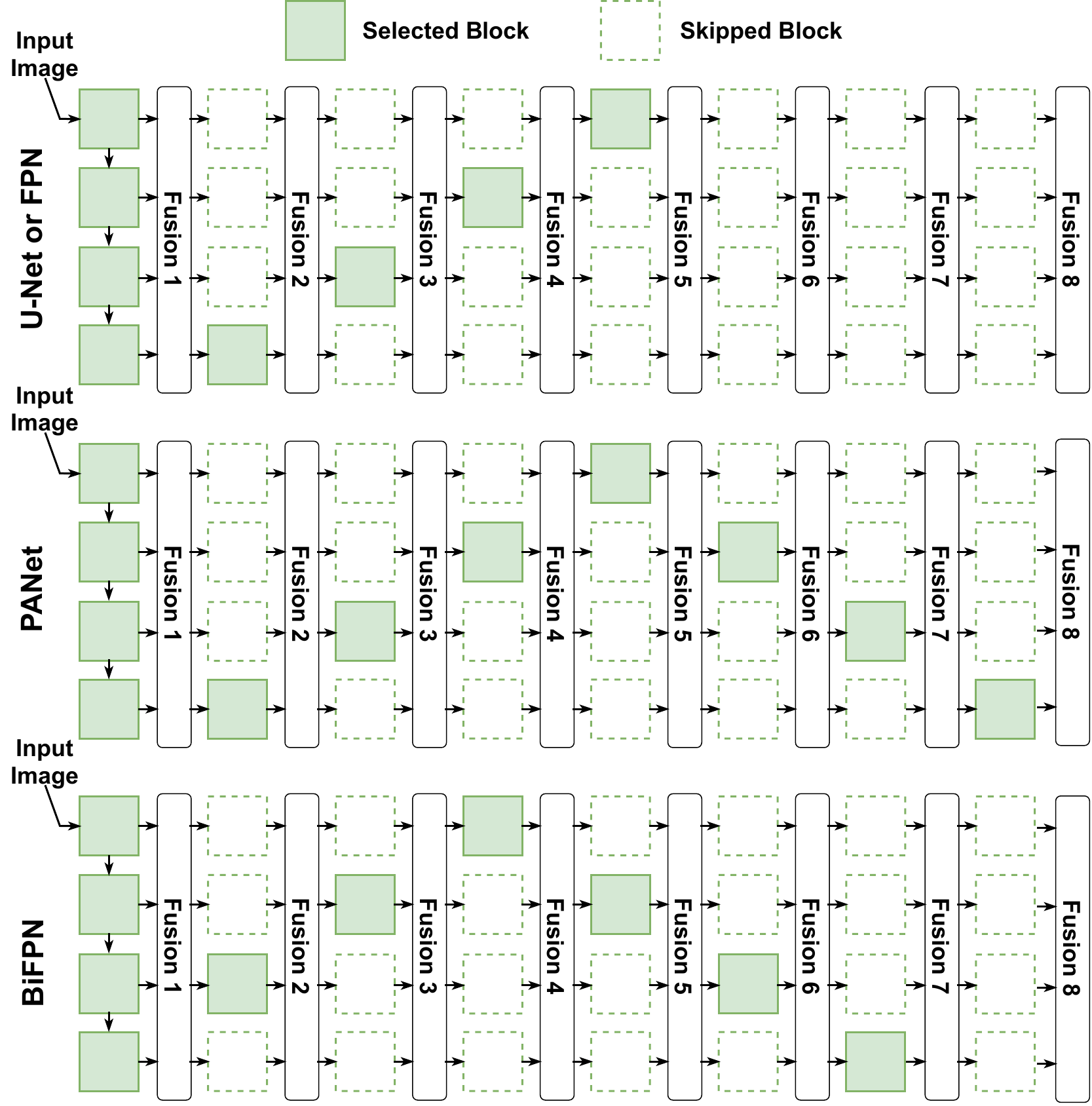}
\vspace{-1.5em}
\caption{The search space can represent the topology of U-Net~\cite{ronneberger2015u}, PANet~\cite{liu2018path}, FPN~\cite{lin2017feature}, and BiFPN (without the extra edge)~\cite{tan2020efficientdet}.}
\label{fig:special_cases}
\end{figure}

To demonstrate our search space is inclusive, we visualize how can it represent some of the mainstream network topologies for CV tasks in Figure~\ref{fig:special_cases}. These include 1) the U-Net (Figure~\ref{fig:special_cases}-top) and PANet (Figure~\ref{fig:special_cases}-middle) topology for semantic segmentation and 2) the FPN (Figure~\ref{fig:special_cases}-top) and BiFPN (Figure~\ref{fig:special_cases}-bottom) topology for object detection.

\section{FBNetV3-A$_C$ and FBNetV3-A$_R$}
\label{sec:shrunken_fbnetv3_a}
We provide the code snippets below to demonstrate the details about the architectures of FBNetV3-A$_C$ and FBNetV3-A$_R$, by mainly shrinking the resolution and channel sizes from FBNetV3-A, respectively.

\begin{lstlisting}[language=Python, caption=Code snippets of FBNetV3-A$_C$]
# FBNetV3-A_C
# input_size: 224
# [[Operator, Channels, Stride, Repeats]]
# Partition 0
[["conv_k3_hs",12,2,1]],
[["ir_k3_hs",12,1,2,{"expansion": 1}]],
[
  ["ir_k3_hs",18,2,1,{"expansion": 3}],
  ["ir_k3_hs",18,1,2,{"expansion": 2}],
],
# Partition 1
[
  ["ir_k3_sehsig_hs",30,2,1,{"expansion": 4}],
  ["ir_k3_sehsig_hs",30,1,3,{"expansion": 2}],
],
# Partition 2
[
  ["ir_k3_hs",54,2,1,{"expansion": 4}],
  ["ir_k3_hs",54,1,3,{"expansion": 2}],
  ["ir_k3_sehsig_hs",80,1,1,{"expansion": 4}],
  ["ir_k3_sehsig_hs",80,1,3,{"expansion": 2}],
],
# Partition 3
[
  ["ir_k3_sehsig_hs",138,2,1,{"expansion": 4}],
  ["ir_k3_sehsig_hs",138,1,3,{"expansion": 3}],
  ["ir_k3_sehsig_hs",168,1,1,{"expansion": 4}],
],
\end{lstlisting}

\begin{lstlisting}[language=Python, caption=Code snippets of FBNetV3-A$_R$]
# FBNetV3-A_R
# input_size: 160
# [[Operator, Channels, Stride, Repeats]]
# Partition 0
[["conv_k3_hs",12,2,1]],
[["ir_k3_hs",12,1,2,{"expansion": 1}]],
[
  ["ir_k3_hs",18,2,1,{"expansion": 4}],
  ["ir_k3_hs",18,1,3,{"expansion": 2}],
],
# Partition 1
[
  ["ir_k3_sehsig_hs",30,2,1,{"expansion": 5}],
  ["ir_k3_sehsig_hs",30,1,4,{"expansion": 3}],
],
# Partition 2
[
  ["ir_k3_hs",54,2,1,{"expansion": 5}],
  ["ir_k3_hs",54,1,4,{"expansion": 3}],
  ["ir_k3_sehsig_hs",80,1,1,{"expansion": 5}],
  ["ir_k3_sehsig_hs",80,1,5,{"expansion": 3}],
],
# Partition 3
[
  ["ir_k3_sehsig_hs",138,2,1,{"expansion": 6}],
  ["ir_k3_sehsig_hs",138,1,5,{"expansion": 4}],
  ["ir_k3_sehsig_hs",168,1,1,{"expansion": 6}],
],
\end{lstlisting}

\section{Important Implementation Details of Algorithm \ref{alg:multitask_supernet_is_reinforce}}
\label{sec:alg_details}
We provide several impotant implementation details of Algorithm \ref{alg:multitask_supernet_is_reinforce}.

\textbf{Sampling multiple architectures}. Algorithm \ref{alg:singletask_supernet} \ref{alg:multitask_supernet}, \ref{alg:multitask_supernet_is}, \ref{alg:multitask_supernet_is_reinforce} show we sample 1 architecture in each forward pass. Although it still gives an unbiased estimation of the task loss, small sample sizes lead to large variations. In practice, we implement the supernet training with \textit{distributed data parallel} in Pytorch, such that each thread independently samples an architecture from the same distribution. We use 16 threads for supernet training, therefore, sampling 16 architectures per iteration to reduce the estimation variance. 

\textbf{Self-normalized importance sampling.} In Equation (\ref{eqn:importance_sampling}), we compute the importance weight as $r(\va) = p_{\vpi}(\va) / q(\va)$. In some extreme cases if $q(\va)$ is too small relative to $ p_{\vpi}(\va)$, $r(\va)$ will become very large that destabilize the supernet training. To prevent this, we actually use the \textit{self-normalized importance sampling} and re-write Equation (\ref{eqn:importance_sampling}) as
\begin{equation}
    \mathbb{E}_{\va^t \sim p_{\vpi^t}}\{\ell^t(\va^t, \vw)\}      \approx  \frac{\sum_{i=1}^N r(\va_i) \ell^t(\va_i, \vw) }{\sum_{i=1}^N r(\va_i)},
\end{equation}
with $\va_i \sim p_{\vpi^t}$.
This still gives an unbiased estimation \cite{mcbook}, but will prevent the loss from becoming exceedingly large. During supernet training, we implement this through an all-gather operation to collect $r(\va_i)$ from all threads and compute the normalized importance weight. 

\textbf{Loss normalization}. In Equation (\ref{eqn:reinforce}), we scale the gradient $\nabla_{\vpi^t} \log p_{\vpi^t}(\va)$ by the associated loss $\ell^t(\va)$ to determine whether we should suppress or encourage the sampled architecture $\va$.  However, one challenge is that for different tasks, the loss $\ell^t$ may have different mean and variance, so the gradients of different tasks can be scaled differently. To address this, instead of using the raw task-loss in Equation (\ref{eqn:reinforce}), we use a normalized task-loss, computed as $\hat{\ell}^t(\va) = (\ell(\va) - \mu_\ell) / \sigma_\ell $, where $\mu_\ell, \sigma_\ell$ is the mean and standard deviation of the task loss in the past 200 steps. The mean $\mu_\ell$ provides a baseline to evaluate how does the sampled architecture $\va$ compare with the average. This is similar to the Reinforcement Learning approach of using ``advantage'' instead of reward for policy gradient. The scaling factor $1/\sigma_\ell$ ensures that all losses are scaled properly without needing to tune the task-specific learning rate. 

\textbf{Cost regularization.} In addition to the original task loss, \eg, cross-entropy for classification, we add a cost regularization term computed as $\lambda_c \max(0, \frac{\sum_b a_b c_b}{\sum_b c_b} - 0.5)$, where $c_b$ is the cost (\eg, FLOPs) of block-$b$, and $a_b$ denotes whether to select block-b or not. $\lambda_c$ is a loss coefficient. $0.5$ is the relative cost target. 

\textbf{Warmup training.} Similar to the observation of \cite{wu2019fbnet}, before training the architecture parameters $\vpi$, we need to first sufficiently train the model weights $\vw$. This is because at the beginning of the supernet training, the loss will always drop regardless of the choice of architecture $\va$. 
In our implementation, we use warmup training to first train $\vw$ sufficiently and then begin to update $\vpi$ and $\vw$ jointly.

\section{Details about the Search and Training Process Implementation}

\subsection{Search Process Implementation}
\label{sec:search_details}
We introduce the implementation details of the search process of {\FrameworkName}. As discussed in Section \ref{sec:method}, our search is conducted by training a supernet on a multitask proxy dataset. Details of the dataset creation can be found in Section \ref{sec:search}, supernet design can be found in Section \ref{sec:space}. Our search is based on the supernet extended from an FBNetV3-A model as illustrated in Section~\ref{sec:space}. On top of the supernet we use the FBNet-V3 style classification head attached to the end of \texttt{Path 3}. We use a Faster R-CNN head attached to \texttt{Path 2} for object detection, and a single convolutional layer as the segmentation head attached to \texttt{Path 2} for semantic segmentation. Note since we only care about topologies, heads used during search can be different from the heads for downstream task. We pretrain the extended supernet on the ImageNet for classification, and then train it on the multitask proxy dataset. We implement the search algorithm in D2go\footnote{https://github.com/facebookresearch/d2go} powered by Pytorch~\cite{paszke2019pytorch} and Detectron2~\cite{wu2019detectron2}. To train the supernet, we use a total batch size of 768. The images are resized such that the short size is 256, and we take a random crop with size 224x224 to feed to the model. We train the supernet for 9375 steps using SGD with a base learning rate of 0.96. We decay the learning rate by 10x at 3125 steps. We set the initial sampling probability of all blocks to 0.5. We do not update the architecture parameters until 6250 steps, and we the architecture parameter's learning rate is 0.01 of the regular learning rate for weights. We use 16 V100 GPUs to train the supernet. It takes about 10 hours to finish. 

\subsection{Training Process Implementation}
\label{sec:train_details}
For training the task-specific architectures searched by {\FrameworkName}, we follow existing SotA training recipes for each task~\cite{graham2021levit,cheng2021maskformer,wu2019detectron2} and use PyTorch~\cite{paszke2019pytorch} for all the experiments. 

For \underline{ImageNet}~\cite{deng2009imagenet} image classification, we use the FBNetV3 style MBPool+FC classification head on top of the final feature map from \texttt{Path 3} in Figure~\ref{fig:overview}. We adopt the distillation based training settings in~\cite{graham2021levit,touvron2020training} and use a large pretrained model that has a 85.5\% top-1 accuracy on ImageNet as the teacher model. We use a batch size of 4096 on 64 V100 GPUs for 1100 epochs, using SGD with momentum 0.9 and weight decay 2$\times 10^{-5}$ as the optimizer, initializing the learning rate as 4.0 with 11 epochs warm-up from 0.01, and decaying it each epoch with a factor of 0.9875.

For \underline{ADE20K} semantic segmentation, we modify the MaskFormer's~\cite{cheng2021maskformer} segmentation head to a lighter version, \ie, we use a pixel decoder with a 3$\times$3 convolution layer and shrink the transformer decoder to only contain 1 Transformer layer, dubbed as Lite MaskFormer. The pixel decoder is attached to the end of \texttt{Path 1}, and the transformer decoder is attached to \texttt{Path 3}. We use the same training settings for ResNet backbone in~\cite{cheng2021maskformer} to train all the searched architectures except using 320k iterations for bigger models ({\FrameworkName}-A/C/F-SEG in Table~\ref{tab:ade20k}) following~\cite{wang2021pyramid}. We initialize the backbone with the weights of the ImageNet-pretrained supernet.

For \underline{COCO}~\cite{lin2014microsoft} object detection, we use the searched architectures as the backbone feature extractor. We attach a Faster R-CNN ~\cite{ren2015faster} head on \texttt{Path 1} of the supernet. We re-design the ROI and RPN head to have a lighter architecture, and reduce the number of ROI proposals to 30 and name this version as Lite R-CNN. We follow most of the default training settings in~\cite{wu2019detectron2} while using a batch size of 256 to train all the searched architectures for 150k iterations with a base learning rate of 0.16, and decay the learning rate by 10 after 140K steps. We keep an exponential moving average (EMA) of the model weights, and evaluate on the EMA model. The same as the settings in ADE20K above, we initialize the backbone with the weights from the ImageNet-pretrained supernet. 

\section{Design of light Detection and Segmentation Head}
\label{sec:head_arch}
\subsection{Architecture of the Lite MaskFormer Head}
\label{sec:lite_maskformer_head}
MaskFormer~\cite{cheng2021maskformer} consists of three components, a pixel decoder (PD), a transformer decoder (TD), and a segmentation module (SM). The pixel decoder is used to generate the per-pixel embeddings. The transformer decoder is designed to output the per-segment embeddings which encode the global information of each segment. The segmentation module converts the per-segment embeddings to mask embeddings via a Multi-Layer Perceptron (MLP), and then obtain final predication via a dot product between the per-pixel embeddings from the pixel decoder and the mask embeddings.

We squeeze both the pixel decoder and the transformer decoder to build the Lite MaskFormer used in our experiments.

Our pixel decoder takes the output from \texttt{Path 1} and leverage a 3$\times$3 convolution layer to generate the per-pixel embeddings.

Our transformer decoder follows the design of MaskFormer's transformer decoder, \ie, the same with DETR~\cite{carion2020end}. But we shrink it to only contain 1 Transformer~\cite{bello2019attention} layer and attach it to the output of \texttt{Path 3}.

We further demonstrate the distribution of our models' FLOPs in Table~\ref{tab:set_flop}. The total FLOPs is the sum of \texttt{BB}, \texttt{PD}, \texttt{TD}, and \texttt{SM} FLOPs, and it is computed based on the input resolution of \texttt{(short\_size $\times$ short\_size)} following~\cite{cheng2021maskformer,xie2021segformer}.

\begin{table}[t]
\caption{FLOPs of {\FrameworkName} segmentation models. \texttt{BB, PD, TD, SM} columns reports the million (M) FLOPs of the backbone, pixel decoder, transformer decoder and segmentation module of a model given the input size as \texttt{(short\_size $\times$ short\_size)}. Column \texttt{Total} is the sum of \texttt{BB, PD, TD, SM}.}
\label{tab:set_flop}

\resizebox{\linewidth}{!}{
\begin{tabular}{ll||lllll}
\toprule
Model & Short Size & \texttt{BB} & \texttt{PD}  & \texttt{TD} & \texttt{SM} & \texttt{Total} \\
\midrule
{\FrameworkName}-A$_C$-SEG & 384 & 945 & 162 & 139 & 82 & 1328\\
{\FrameworkName}-A$_R$-SEG & 384 & 1389 & 162  & 139 & 82 & 1773 \\
{\FrameworkName}-A-SEG & 384 & 2485 & 215 & 135 & 84 & 2919 \\
{\FrameworkName}-C-SEG & 448 & 3838 & 357  & 144 & 109 & 4448  \\
{\FrameworkName}-F-SEG & 512 & 8502 & 550 & 155 & 142 & 9350 \\
\bottomrule
\end{tabular}
}
\end{table}

\subsection{Architecture of the Lite Faster R-CNN Head}
\label{sec:lite_rcnn_head}

\begin{table}[b]
\caption{FLOPs of {\FrameworkName} detection models. \texttt{BB, RPN, ROI} columns reports the million (M) FLOPs of the backbone, RPN, and ROI of a model given the reference input size. Column \texttt{Total} is the sum of \texttt{BB, RPN, ROI}. Column \texttt{Avg.} reports the average FLOPs of the model on the COCO \textit{val} set.   }
\label{tab:det_flop}
\resizebox{\linewidth}{!}{
\begin{tabular}{ll||lllll}
\toprule
Model                      & Ref. Size & \texttt{BB} & \texttt{RPN}  & \texttt{ROI} & \texttt{Total} & \texttt{Avg.} \\
\midrule
{\FrameworkName}-A$_C$-DET & 213x320         & 399      & 152  & 182 & 733  & 713   \\
{\FrameworkName}-A$_R$-DET & 213x320         & 601      & 152  & 182 & 935  & 908   \\
{\FrameworkName}-A-DET     & 213x320         & 1054     & 158  & 186 & 1398 & 1354  \\
{\FrameworkName}-A$_C$-DET & 320x481         & 912      & 347  & 182 & 1441 & 1367  \\
{\FrameworkName}-A$_R$-DET & 320x481         & 1372     & 347  & 182 & 1901 & 1800  \\ 
\bottomrule
\end{tabular}
}
\end{table}

For object detection, we attach Faster R-CNN~\cite{ren2015faster} head to our searched backbones. Faster R-CNN detection contains two component, a region proposal network (RPN) and a region-of-interest (ROI) head. We use light-weight RPN and ROI heads to save the overall compute cost. 

Our ROI head contains a inverted resitual block (IRB) \cite{sandler2018mobilenetv2} with kernel size 3, expansion ratio 3, output channel size 96. We also use Squeeze-Excitation \cite{hu2018squeeze} and HSigmoid activation following \cite{howard2019searching}. The output of RPN is fed to a single convolution layer to generate RPN output. 

Our RPN head contains 4 IRB blocks with the same kernel size of 3; expansion ration of 4, 6, 6, 6; output channel size of 128, 128, 128, 160. The IRB blocks do not use SE or HSigmoid. We use an \texttt{ROIPool} operator to extract feature maps from a region-of-interest, and reshape the spatial size to 6x6. The first IRB block further down-samples the input resolution to 3x3. The output of the IRB blocks are projected by a single conv layer to predict ROI output (bounding box prediction, class prediction, \textit{etc.}). 

During inference, we select the 30 regions post NMS and feed them to ROI. Under this setting, our models FLOPs distribution is shown in Table \ref{tab:det_flop}.  Note that the total FLOPs of our model is computed based on the reference input size. The total FLOPs is the sum of \texttt{BB}, \texttt{RPN}, and \texttt{ROI} FLOPs. The average FLOPs reported in Table \ref{tab:coco} is computed based on images in the COCO \textit{val} set.

\section{Average FLOPs of R-CNN models.}
\label{sec:avg_flop}
In Table \ref{tab:coco} and Table \ref{tab:det_flop}, we report the \textit{average} FLOPs of our model on the COCO validation dataset. This is because our R-CNN based detection model does not fix the input size, while our baselines \cite{nanodet,ge2021yolox} takes a fixed input size. It is a more fair to use the average FLOPs of R-CNN models to compare models with a fixed input size. 

During inference, our R-CNN model re-size images using the following strategy. We first define two parameters \texttt{min\_size} (set to 224 or 320) and \texttt{max\_size} (set to 320 or 640). For an input image, we first resize the image such that its short size becomes \texttt{min\_size} while keeping the aspect ratio the same. After this, if the longer side the image becomes larger than \texttt{max\_size}, we re-size the image again to make sure the longer side becomes \texttt{max\_size}, while not changing the aspect ratio. 

To compute the average FLOPs, we first compute the backbone (BB), RPN, ROI, and total flops of the model based on a reference input (\eg, 213x320 or 320x481), as in Table \ref{tab:det_flop}. Then, we compute the number of pixels in the reference image, and the average number of pixels for all images in the dataset. We compute a ratio \texttt{ratio} between the average and reference pixel number. Finally, we compute the average FLOPs as \texttt{ratio x (BB + RPN) + ROI}, where \texttt{BB},  \texttt{RPN},  \texttt{ROI} denotes the backbone, RPN, ROI FLOPs of the model. We do not scale \texttt{ROI} since the backbone and RPN flops is determined by the input resolution while ROI's FLOPs do not depend on input resolution. 

\end{document}